\documentclass[11pt,a4paper]{article}
\usepackage[hyperref]{emnlp-ijcnlp-2019}
\usepackage{times}
\usepackage{amsmath,graphicx}
\usepackage{latexsym}
\usepackage{CJKutf8}
\usepackage[T1]{fontenc}
\usepackage{multirow}
\usepackage{amssymb}
\usepackage{subfig}
\usepackage{flexisym}
\usepackage{url}
\usepackage{caption}

\usepackage{url}

\aclfinalcopy % Uncomment this line for the final submission

\title{Hierarchical Meta-Embeddings for Code-Switching \\Named Entity Recognition}
\author{Genta Indra Winata, Zhaojiang Lin, Jamin Shin, Zihan Liu, Pascale Fung \\
Center for Artificial Intelligence Research (CAiRE)\\
  Department of Electronic and Computer Engineering\\
  The Hong Kong University of Science and Technology, Clear Water Bay, Hong Kong\\
  \texttt{\{giwinata,zlinao,jmshinaa,zliucr\}@connect.ust.hk},\\ \texttt{ pascale@ece.ust.hk}}

\date{}

\begin{document}
\maketitle
\begin{abstract}
  In countries that speak multiple main languages, mixing up different languages within a conversation is commonly called \textit{code-switching}. Previous works addressing this challenge mainly focused on word-level aspects such as word embeddings. However, in many cases, languages share common subwords, especially for closely related languages, but also for languages that are seemingly irrelevant. Therefore, we propose \textit{Hierarchical Meta-Embeddings} (HME) that learn to combine multiple monolingual word-level and subword-level embeddings to create language-agnostic lexical representations. On the task of Named Entity Recognition for English-Spanish code-switching data, our model achieves the state-of-the-art performance in the multilingual settings. We also show that, in cross-lingual settings, our model not only leverages closely related languages, but also learns from languages with different roots. Finally, we show that combining different subunits are crucial for capturing code-switching entities. 
  %Finally, we show that subword-level modeling is crucial for capturing code-switching entities.

\end{abstract}

\section{Introduction}
% Each paragraph should answer the following questions:
% P1. What is code switching and why is it important?
% P2. How has this problem been approached so far?
% P3. What are the weakness / limitations / or missing points of the previous works?
% P4. How do you propose to tackle and what are the intuitions?
% P5. What are your results and claims??

% P1. What is code switching and why is it important?
% P2. How has this problem been approached so far?
\textit{Code-switching} is a phenomenon that often happens between multilingual speakers, in which they switch between their two languages in conversations, hence, it is practically useful to recognize this well in spoken language systems \cite{winata2018code}. This occurs more often for entities such as organizations or products, which motivates us to focus on the specific problem of Named Entity Recognition (NER) in code-switching scenarios. We show one of the examples as the following:
\begin{itemize}
  \item[$\bullet$] \textit{walking dead} le quita el apetito a cualquiera
  \item[$\bullet$] \textbf{(translation)} \textit{walking dead} (a movie title) takes away the appetite of anyone
\end{itemize}

For this task, previous works have mostly focused on applying pre-trained word embeddings from each language in order to represent noisy mixed-language texts, and combine them with character-level representations~\cite{trivedi2018iit,wang2018code,winata2018bilingual}.
% P3. What are the weakness / limitations / or missing points of the previous works?
% P4. How do you propose to tackle and what are the intuitions?
However, despite the effectiveness of such word-level approaches, they neglect the importance of subword-level characteristics shared across different languages. Such information is often hard to capture with word embeddings or randomly initialized character-level embeddings. Naturally, we can turn towards subword-level embeddings such as FastText~\cite{grave2018learning} to help this task, which will evidently allow us to leverage the morphological structure shared across different languages.

\begin{figure*}[!t]
    \centering
    \includegraphics[width=0.92\linewidth]{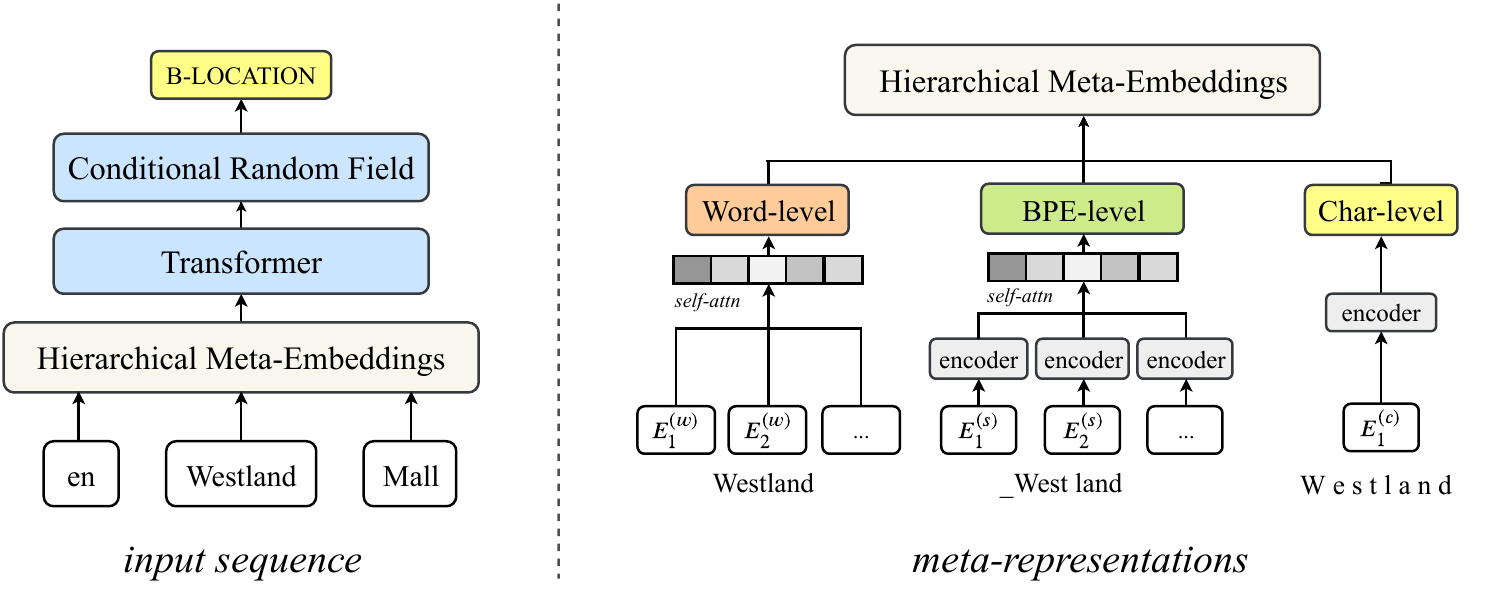}
    \caption{Hierarchical Meta-Embeddings (HME) architecture for Named Entity Recognition (NER) task. \textbf{Left:} Transformer-CRF architecture for Named Entity Recognition. \textbf{Right:} HME accept words, BPEs, and characters inputs.}
    \label{fig:model}
\end{figure*}

Despite such expected usefulness, there has not been much attention focused around using subword-level features in this task. 
This is partly because of the non-trivial difficulty of combining different language embeddings in the subword space, which arises from the distinct segmentation into subwords for different languages. 
This leads us to explore the literature of Meta-Embeddings~\cite{yin2016learning,muromagi2017linear,bollegala2018think,coates2018frustratingly,kiela2018dynamic,winata2019learning}, which is a method to learn how to combine different embeddings. 

% P4. How do you propose to tackle and what are the intuitions?
In this paper, we propose \textbf{Hierarchical Meta-Embeddings (HME)}~\footnote{The source code is available at \url{https://github.com/gentaiscool/meta-emb}} which learns how to combine different pre-trained monolingual embeddings in word, subword, and character-level into a single language-agnostic lexical representation without using specific language identifiers. To address the issue of different segmentations, we add a Transformer~\cite{vaswani2017attention} encoder which learns the important subwords in a given sentence. We evaluate our model on the task of Named Entity Recognition for English-Spanish code-switching data, and we use  \textbf{Transformer-CRF}, a transformer-based encoder for sequence labeling based on the implementation of \citet{winata2019learning}.
% P5. What are your results and claims??
Our experimental results confirm that HME significantly outperforms the state-of-the-art system in \textit{absolute F1 score}. The analysis shows that in the task of English-Spanish mixed texts not only similar languages like Portuguese or Catalan help, but also seemingly distant languages from Celtic origin also significantly increase the performance.

\section{Related Work}

\paragraph{Embeddings}
Previous works have extensively explored different representations such as word \cite{mikolov2013distributed,pennington2014glove,grave2018learning,xu2018emo2vec}, subword \cite{sennrich2016neural,heinzerling2018bpemb}, and character \cite{Santos2014LearningCR,wieting2016charagram}. 
\citet{lample2016neural} has successfully concatenated character and word embeddings to their model, showing the potential of combining multiple representations. \citet{liu-etal-2019-incorporating} proposed to leverage word and subword embeddings into the application of unsupervised machine translation.

\paragraph{Meta-embeddings}
Recently, there are studies on combining multiple word embeddings in pre-processing steps  \cite{yin2016learning,muromagi2017linear,bollegala2018think,coates2018frustratingly}. Later, \citet{kiela2018dynamic} introduced a method to dynamically learn word-level meta-embeddings, which can be effectively used in a supervised setting. \citet{winata2019learning} proposed an idea to leverage multiple embeddings from different languages to generate language-agnostic meta-representations for mixed-language data. 
% In another line of work, the idea of combining embeddings is related to multi-modal learning \cite{kiela2014learning,lazaridou2015combining,Wang2018AssociativeMA,wang2018learning}.

\section{Hierarchical Meta-Embeddings}
We propose a method to combine word, subword, and character representations to create a mixture of embeddings. We generate a multilingual meta-embeddings of word and subword, and then, we concatenate them with character-level embeddings to generate final word representations, as shown in Figure~\ref{fig:model}. Let $\mathbf{w}$ be a sequence of words with $n$ elements, where $\mathbf{w} = [w_1,\dots,w_n]$. Each word can be tokenized into a list of subwords $\mathbf{s} = [s_1,\dots,s_m]$ and a list of characters $\mathbf{c} = [c_1,\dots,c_p]$. The list of subwords $\mathbf{s}$ is generated using a function $f$; $\mathbf{s} = f(\mathbf{w})$. Function $f$ maps a word into a sequence of subwords. Further, let $E^{(w)}$, $E^{(s)}$, and $E^{(c)}$ be a set of word, subword, and character embedding lookup tables. Each set consists of different monolingual embeddings. Each element is transformed into a embedding vector in $\mathbb{R}^d$. We denote subscripts $_{\{i,j\}}$ as element and embedding language index, and superscripts $^{({w,s,c})}$ as word, subword, and character.

\subsection{Multilingual Meta-Embeddings (MME)}
We generate a meta-representations by taking the vector representation from multiple monolingual pre-trained embeddings in different subunits such as word and subword. We apply a \textbf{projection matrix} $\mathbf{W}_j$ to transform the dimensions from the original space $\mathbf{x}_{i,j} \in \mathbb{R}^d$ to a new shared space $\mathbf{x\textprime}_{i,j} \in \mathbb{R}^{d'}$. Then, we calculate \textbf{attention weights} $\alpha_{i,j} \in \mathbb{R}^{d'}$ with a non-linear scoring function $\phi$ (e.g., tanh) to take important information from each individual embedding $\mathbf{x\textprime}_{i,j}$. Then, MME is calculated by taking the weighted sum of the projected embeddings $\mathbf{x\textprime}_{i,j}$:
\begin{align}
\mathbf{x\textprime}_{i,j} &= \mathbf{W}_j \cdot \mathbf{x}_{i,j}, \\
\alpha_{i,j} &= \frac{\exp(\phi(\mathbf{x\textprime}_{i,j}))}{\sum_{k=1}^n  \exp(\phi(\mathbf{x\textprime}_{i,k}))}, \\
\mathbf{u}_i &= \sum_{j=1}^n {\alpha_{i,j} 
\mathbf{x\textprime}_{i,j}}.
% \textbf{w\textprime}_{i,j} &= \textbf{a}_j \cdot \textbf{w}_{i,j} + b_j
\end{align}
% where $\phi$ is a tanh function.

\begin{table*}[!t]
\centering
\resizebox{\textwidth}{!}{
\begin{tabular}{llllllll}
\hline
\multicolumn{1}{c|}{\multirow{4}{*}{Model}} & \multicolumn{4}{c|}{Multilingual embeddings} & \multicolumn{3}{c}{Cross-lingual  embeddings} \\ \cline{2-8} 
\multicolumn{1}{c|}{} & \multicolumn{1}{c|}{\begin{tabular}[c]{@{}c@{}}main\\ languages\end{tabular}} & \multicolumn{2}{c|}{\begin{tabular}[c]{@{}c@{}}+ closely-related\\ languages\end{tabular}} & \multicolumn{1}{c|}{\begin{tabular}[c]{@{}c@{}}+ distant\\ languages\end{tabular}} & \multicolumn{2}{c|}{\begin{tabular}[c]{@{}c@{}}closely-related\\ languages\end{tabular}} & \multicolumn{1}{c}{\begin{tabular}[c]{@{}c@{}}distant\\ languages\end{tabular}} \\ \cline{2-8} 
\multicolumn{1}{c|}{} & \multicolumn{1}{c|}{en-es} & \multicolumn{1}{c|}{ca-pt} & \multicolumn{1}{c|}{ca-pt-de-fr-it} & \multicolumn{1}{c|}{br-cy-ga-gd-gv} & \multicolumn{1}{c|}{ca-pt} & \multicolumn{1}{c|}{ca-pt-de-fr-it} & br-cy-ga-gd-gv \\ \hline \hline
\multicolumn{8}{l}{\textit{Flat word-level embeddings}} \\ \hline
\multicolumn{1}{l|}{CONCAT} &  \multicolumn{1}{c|}{65.3 \small{$\pm$ 0.38}} & \multicolumn{1}{c|}{64.99 \small{$\pm$ 1.06}} & \multicolumn{1}{c|}{65.91 \small{$\pm$ 1.16}} & \multicolumn{1}{c|}{65.79 \small{$\pm$ 1.36}} & \multicolumn{1}{c|}{58.28 \small{$\pm$ 2.66}} & \multicolumn{1}{c|}{64.02 \small{$\pm$ 0.26}} & \multicolumn{1}{c}{50.77 \small{$\pm$ 1.55}} \\
\multicolumn{1}{l|}{LINEAR} & \multicolumn{1}{c|}{64.61 \small{$\pm$ 0.77}} & \multicolumn{1}{c|}{65.33 \small{$\pm$ 0.87}} & \multicolumn{1}{c|}{65.63 \small{$\pm$ 0.92}} & \multicolumn{1}{c|}{64.95 \small{$\pm$ 0.77}} & \multicolumn{1}{c|}{60.72 \small{$\pm$ 0.84}} & \multicolumn{1}{c|}{62.37 \small{$\pm$ 1.01}} & \multicolumn{1}{c}{53.28 \small{$\pm$ 0.41}}\\ \hline
\multicolumn{8}{l}{\textit{Multilingual Meta-Embeddings (MME)} \cite{winata2019learning}} \\ \hline
\multicolumn{1}{l|}{Word} &
\multicolumn{1}{c|}{65.43 \small{$\pm$ 0.67}} & \multicolumn{1}{c|}{66.63 \small{$\pm$ 0.94}} & 
\multicolumn{1}{c|}{66.8 \small{$\pm$ 0.43}} & \multicolumn{1}{c|}{66.56 \small{$\pm$ 0.4}} & \multicolumn{1}{c|}{61.75 \small{$\pm$ 0.56}} & \multicolumn{1}{c|}{63.23 \small{$\pm$ 0.29}} & \multicolumn{1}{c}{53.43 \small{$\pm$ 0.37}} \\ \hline  
\multicolumn{8}{l}{\textit{Hierarchical Meta-Embeddings (HME)}} \\ \hline
\multicolumn{1}{l|}{+ BPE} & \multicolumn{1}{c|}{65.9 \small{$\pm$ 0.72}} & \multicolumn{1}{c|}{67.31 \small{$\pm$ 0.34}} & \multicolumn{1}{c|}{67.26 \small{$\pm$ 0.54}} & \multicolumn{1}{c|}{66.88 \small{$\pm$ 0.37}} & \multicolumn{1}{c|}{63.44 \small{$\pm$ 0.33}} & \multicolumn{1}{c|}{63.78 \small{$\pm$ 0.62}} & \multicolumn{1}{c}{60.19 \small{$\pm$ 0.63}}\\
\multicolumn{1}{l|}{+ Char} & \multicolumn{1}{c|}{65.88 \small{$\pm$ 1.02}} & \multicolumn{1}{c|}{67.38 \small{$\pm$ 0.84}} & \multicolumn{1}{c|}{65.94 \small{$\pm$ 0.47}} & 
\multicolumn{1}{c|}{66.1 \small{$\pm$ 0.33}} & 
\multicolumn{1}{c|}{61.97 \small{$\pm$ 0.6}} & 
\multicolumn{1}{c|}{63.06 \small{$\pm$ 0.69}} & 
\multicolumn{1}{c}{57.5 \small{$\pm$ 0.56}}\\
\multicolumn{1}{l|}{+ BPE + Char} & \multicolumn{1}{c|}{66.55 \small{$\pm$ 0.72}} & \multicolumn{1}{c|}{\textbf{67.8 \small{$\pm$ 0.31}}} & \multicolumn{1}{c|}{67.07 \small{$\pm$ 0.49}} & \multicolumn{1}{c|}{67.02 \small{$\pm$ 0.16}} & \multicolumn{1}{c|}{63.9 \small{$\pm$ 0.22}} & \multicolumn{1}{c|}{ \textbf{64.52 \small{$\pm$ 0.35 }}} & \multicolumn{1}{c}{60.88 \small{$\pm$ 0.84}} \\ \hline
% \multicolumn{1}{l|}{only BPE} & \multicolumn{1}{c|}{ 57.73 \small{$\pm$ 0.74 }} &  & & & & & \\ \hline
\end{tabular}
}
\caption{Results (percentage F1 mean and standard deviation from five experiments). \textbf{Multilingual}: with main languages, \textbf{Cross-lingual}: without main languages. }
\label{tab:res}
\end{table*}

\subsection{Mapping Subwords and Characters to Word-Level Representations} 
We propose to map subword into word representations and choose byte-pair encodings (BPEs)~\cite{sennrich2016neural} since it has a compact vocabulary. First, we apply $f$ to segment words into sets of subwords, and then we extract the pre-trained subword embedding vectors $\mathbf{x}^{(s)}_{i,j} \in \mathbb{R}^d$ for language $j$.

Since, each language has a different $f$, we replace the projection matrix with Transformer~\cite{vaswani2017attention} to learn and combine important subwords into a single vector representation. Then, we create $\mathbf{u}_i^{(s)} \in \mathbb{R}^{d'}$ which represents the subword-level MME by taking the weighted sum of $\mathbf{x}\textprime^{(s)}_{i,j} \in \mathbb{R}^{d'}$.
\begin{align}
\mathbf{x}\textprime^{(s)}_{i,j} = \text{Encoder}(\mathbf{x}^{(s)}_{i,j}),\\
\mathbf{u}_{i}^{(s)} = \sum_{j=1}^n{\alpha_{i,j} \mathbf{x}\textprime^{(s)}_{i,j}}.
\end{align}

\begin{table}[!t]
\centering
\resizebox{0.44\textwidth}{!}{
\begin{tabular}{l|c}
\hline
\multicolumn{1}{c|}{Model} & \multicolumn{1}{c}{F1} \\ \hline \hline
\multicolumn{1}{l|}{\citet{trivedi2018iit} } & 61.89 \\
\multicolumn{1}{l|}{\citet{wang2018code}} & 62.39 \\
\multicolumn{1}{l|}{\citet{wang2018code} (Ensemble)} & 62.67 \\
\multicolumn{1}{l|}{\citet{winata2018bilingual}} & 62.76 \\
\multicolumn{1}{l|}{\citet{trivedi2018iit} (Ensemble)} & 63.76 \\
\multicolumn{1}{l|}{\citet{winata2019learning} MME} & 66.63 \small{$\pm$ 0.94} \\ \hline
% \multicolumn{1}{l|}{\citet{winata2019learning} MME (Ensemble)} & 68.34 \\ \hline
Random embeddings & 46.68 \small{$\pm$ 0.79} \\  \hline
\textit{Aligned embeddings} & \\ 
MUSE (es $\rightarrow$ en) & 60.89 \small{$\pm$ 0.37} \\
MUSE (en $\rightarrow$ es) & 61.49 \small{$\pm$ 0.62} \\ \hline
\textit{Multilingual embeddings} & \\
Our approach & 67.8 \small{$\pm$ 0.31} \\ 
Our approach (Ensemble)$^\dagger$ & \textbf{69.17} \\ \hline
\textit{Cross-lingual embeddings} & \\
Our approach & 64.52 \small{$\pm$ 0.35} \\ 
Our approach (Ensemble)$^\dagger$ & \textbf{65.99} \\ \hline
\end{tabular}
}
\caption{Comparison to existing works. \textbf{Ensemble:} We run a majority voting scheme from five different models.}
% \caption{Comparison to existing works. $^\dagger$ Ensemble five different models. Our approaches are best performing ones from Table~\ref{tab:res}.}
\label{tab:existing}
\end{table}

To combine character-level representations, we apply an encoder to each character.
\begin{align}
    \mathbf{u}_i^{(c)} = \textnormal{Encoder}(\mathbf{x}_{i}) \in \mathbb{R}^{d'}.
\end{align}

We combine the word-level, subword-level, and character-level representations by concatenation $\mathbf{u}^{HME}_i = (\mathbf{u}^{(w)}_i, \mathbf{u}^{(s)}_i, \mathbf{u}^{(c)}_i)$, where $\mathbf{u}^{(w)}_i \in \mathbb{R}^{d'}$ and $\mathbf{u}^{(s)}_i \in \mathbb{R}^{d'}$ are word-level MME and BPE-level MME, and $\mathbf{u}^{(c)}_i$ is a character embedding. We randomly initialize the character embedding and keep it trainable. We fix all subword and word pre-trained embeddings during the training.

\subsection{Sequence Labeling}
To predict the entities, we use Transformer-CRF, a transformer-based encoder followed by a Conditional Random Field (CRF) layer \cite{Lafferty2001ConditionalRF}. 
The CRF layer is useful to constraint the dependencies between labels.
\begin{align}
h = \text{Transformer}(\mathbf{u}), h' = \text{CRF}(h).
\end{align}
The best output sequence is selected by a forward propagation using the Viterbi algorithm.

\section{Experiments}
\subsection{Experimental Setup}
We train our model for solving Named Entity Recognition on English-Spanish code-switching tweets data from \citet{W18-3219}. There are nine entity labels with IOB format. The training, development, and testing sets contain 50,757, 832, and 15,634 tweets, respectively. 

We use FastText word embeddings trained from Common Crawl and Wikipedia \cite{grave2018learning} for English (\textit{es}), Spanish (\textit{es}), including \textbf{four Romance languages}: Catalan (\textit{ca}), Portuguese (\textit{pt}), French (\textit{fr}), Italian (\textit{it}), and \textbf{a Germanic language}: German (\textit{de}), and \textbf{five Celtic languages} as the distant language group: Breton (\textit{br}), Welsh (\textit{cy}), Irish (\textit{ga}), Scottish Gaelic (\textit{gd}), Manx (\textit{gv}). We also add the English Twitter GloVe word embeddings \cite{pennington2014glove} and BPE-based subword embeddings from \citet{heinzerling2018bpemb}. 

We train our model in two different settings: \textbf{(1) multilingual setting}, we combine main languages (\textit{en-es}) with Romance languages and a Germanic language, and \textbf{(2) cross-lingual setting}, we use Romance and Germanic languages without main languages. Our model contains four layers of transformer encoders with a hidden size of 200, four heads, and a dropout of 0.1. We use Adam optimizer and start the training with a learning rate of 0.1 and an early stop of 15 iterations. We replace user hashtags and mentions with \texttt{<USR>}, emoji with \texttt{<EMOJI>}, and URL with \texttt{<URL>}. We evaluate our model using \textit{absolute F1 score} metric.

\subsection{Baselines}
% We compare our approach to baselines: 
% Please refer to Appendix for training hyper-parameters details. 
\paragraph{CONCAT} We concatenate word embeddings by merging the dimensions of word representations. This method combines embeddings into a high-dimensional input that may cause inefficient computation.
\begin{equation}
\mathbf{x}_i^{CONCAT} = [\mathbf{x}_{i,1},...,\mathbf{x}_{i,n}].
\end{equation}
\paragraph{LINEAR} We sum all word embeddings into a single word vector with equal weight. This method combines embeddings without considering the importance of each of them.
\begin{equation}
\mathbf{x}_i^{LINEAR} = \sum_{j=0}^n{\mathbf{x\textprime}_{i,j}}.
\end{equation}
\paragraph{Random Embeddings} We use randomly initialized word embeddings and keep it trainable to calculate the lower-bound performance.

\paragraph{Aligned Embeddings} We align English and Spanish FastText embeddings using CSLS with two scenarios. We set English (en) as the source language and Spanish (es) (en $\rightarrow$ es) as the target language, and vice versa (es $\rightarrow$ en). We run MUSE by using the code prepared by the authors of \citet{lample2018word}.~\footnote{https://github.com/facebookresearch/MUSE}

\section{Results \& Discussion}
In general, from Table~\ref{tab:res}, we can see that word-level meta-embeddings even without subword or character-level information, consistently perform better than flat baselines (e.g., CONCAT and LINEAR) in all settings. This is mainly because of the attention layer which does not require additional parameters. Furthermore, comparing our approach to previous state-of-the-art models, we can clearly see that our proposed approaches all significantly outperform them.
% in this task~\cite{aguilar2018named}, we can clearly see that our proposed approaches all significantly outperform them.

From Table~\ref{tab:res}, in Multilingual setting, which trains with the main languages, it is evident that adding both closely-related and distant language embeddings improves the performance. This shows us that our model is able to leverage the lexical similarity between the languages. This is more distinctly shown in Cross-lingual setting as using distant languages significantly perform less than using closely-related ones (e.g., \textit{ca-pt}). Interestingly, for distant languages, when adding subwords, we can still see a drastic performance increase. We hypothesize that even though the characters are mostly different, the lexical structure is similar to our main languages.

\begin{figure*}[!t]
    \centering
    \includegraphics[width=0.98\linewidth]{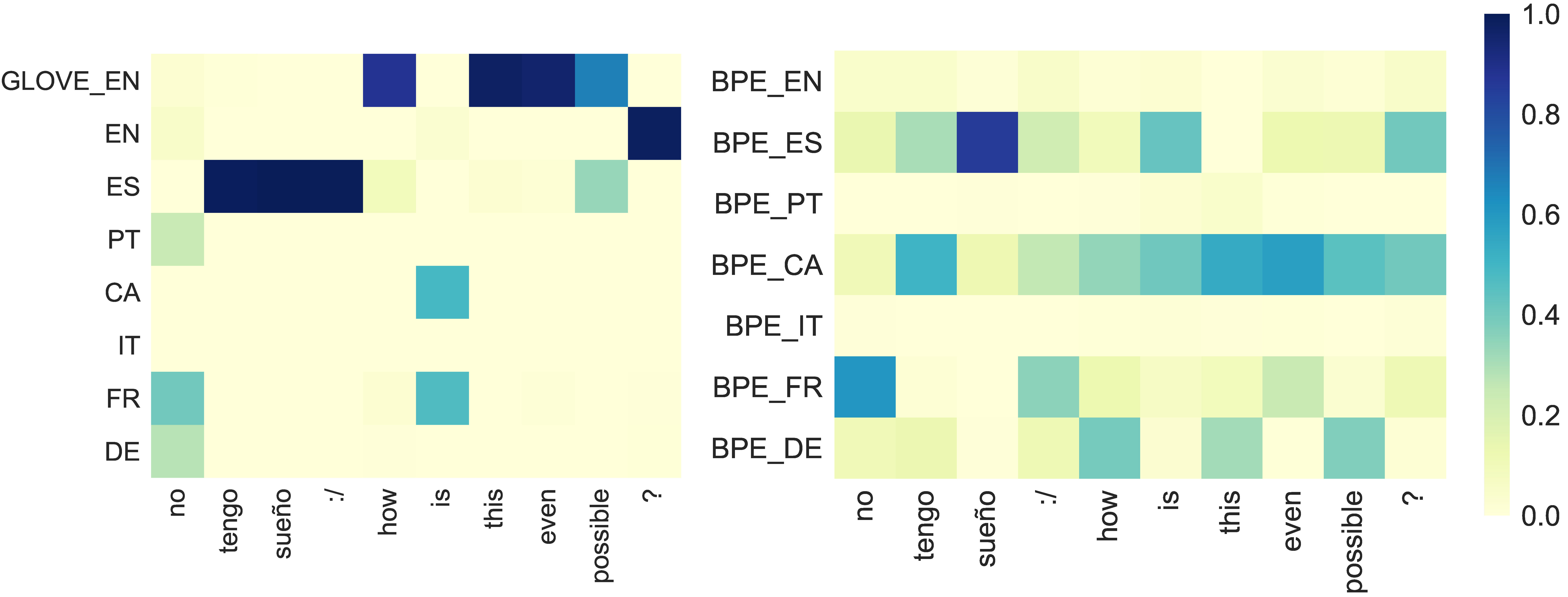}
    \caption{Heatmap of attention over languages from a validation sample. \textbf{Left}: word-level MME, \textbf{Right}: BPE-level MME. We extract the attention weights from a multilingual model (\textit{en-es-ca-pt-de-fr-it}).}
    \label{fig:heatmap}
\end{figure*}

\begin{figure}[!t]
    \centering
    \includegraphics[width=0.98\linewidth]{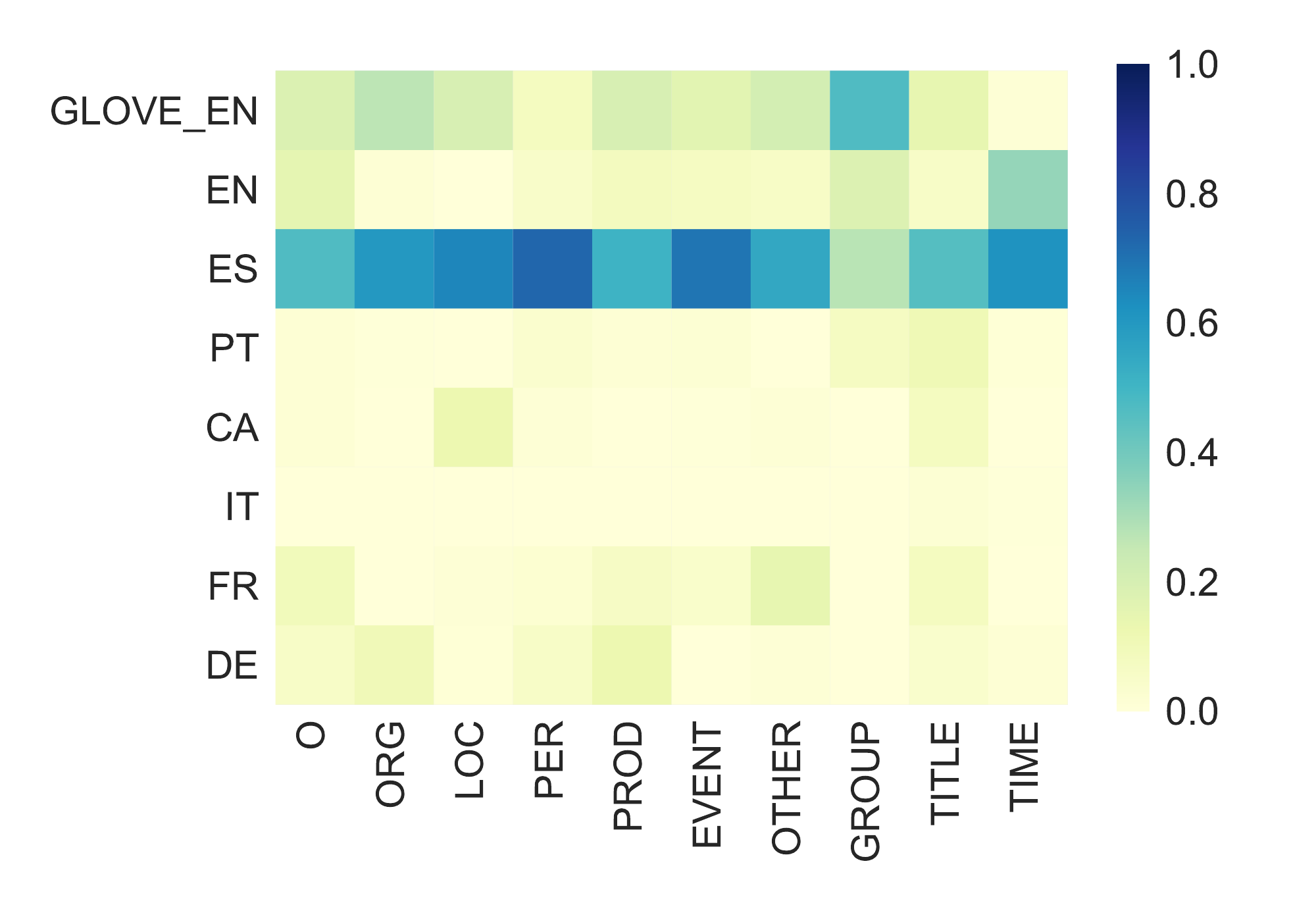}
    \caption{The average of attention weights for word embeddings versus NER tags from the validation set.}
    \label{fig:lang_class}
\end{figure}

On the other hand, adding subword inputs to the model is consistently better than characters. This is due to the transfer of the information from the pre-trained subword embeddings. As shown in Table~\ref{tab:res}, subword embeddings is more effective for distant languages (Celtic languages) than closely-related languages such as Catalan or Portuguese.

Moreover, we visualize the attention weights of the model in word and subword-level to interpret the model dynamics. From the left image of Figure~\ref{fig:heatmap}, in word-level, the model mostly chooses the correct language embedding for each word, but also combines with different languages. Without any language identifiers, it is impressive to see that our model learns to attend to the right languages. The right side of Figure~\ref{fig:heatmap}, which shows attention weight distributions for subword-level, demonstrates interesting behaviors, in which for most English subwords, the model leverages \textit{ca}, \textit{fr}, and \textit{de} embeddings. We hypothesize this is because the dataset is mainly constructed with Spanish words, which can also be verified from Figure~\ref{fig:lang_class} in which most NER tags are classified as \textit{es}.

\section{Conclusion}
We propose \textit{Hierarchical Meta-Embeddings} (HME) that learns how to combine multiple monolingual word-level and subword-level embeddings to create language-agnostic representations without specific language information. We achieve the state-of-the-art results on the task of Named Entity Recognition for English-Spanish code-switching data. We also show that our model can leverage subword information very effectively from languages from different roots to generate better word representations. 

\section*{Acknowledgments}
This work has been partially funded by ITF/319/16FP and MRP/055/18 of the Innovation Technology Commission, the Hong Kong SAR Government, and School of Engineering Ph.D. Fellowship Award, the Hong Kong University of Science and Technology, and RDC 1718050-0 of EMOS.AI. We sincerely thank the three anonymous reviewers for their insightful comments on our paper.

% The acknowledgments should go immediately before the references.  Do
% not number the acknowledgments section. Do not include this section
% when submitting your paper for review. 
\bibliography{emnlp-ijcnlp-2019}

\begin{thebibliography}{24}
\expandafter\ifx\csname natexlab\endcsname\relax\def\natexlab#1{#1}\fi

\bibitem[{Aguilar et~al.(2018)Aguilar, AlGhamdi, Soto, Diab, Hirschberg, and
  Solorio}]{W18-3219}
Gustavo Aguilar, Fahad AlGhamdi, Victor Soto, Mona Diab, Julia Hirschberg, and
  Thamar Solorio. 2018.
\newblock \href {https://www.aclweb.org/anthology/W18-3219} {Named entity
  recognition on code-switched data: Overview of the calcs 2018 shared task}.
\newblock In \emph{Proceedings of the Third Workshop on Computational
  Approaches to Linguistic Code-Switching}, pages 138--147, Melbourne,
  Australia. Association for Computational Linguistics.

\bibitem[{Bollegala et~al.(2018)Bollegala, Hayashi, and
  Kawarabayashi}]{bollegala2018think}
Danushka Bollegala, Kohei Hayashi, and Ken-Ichi Kawarabayashi. 2018.
\newblock Think globally, embed locally: locally linear meta-embedding of
  words.
\newblock In \emph{Proceedings of the 27th International Joint Conference on
  Artificial Intelligence}, pages 3970--3976. AAAI Press.

\bibitem[{Coates and Bollegala(2018)}]{coates2018frustratingly}
Joshua Coates and Danushka Bollegala. 2018.
\newblock Frustratingly easy meta-embedding--computing meta-embeddings by
  averaging source word embeddings.
\newblock In \emph{Proceedings of the 2018 Conference of the North American
  Chapter of the Association for Computational Linguistics: Human Language
  Technologies, Volume 2 (Short Papers)}, pages 194--198.

\bibitem[{Conneau et~al.(2017)Conneau, Lample, Ranzato, Denoyer, and
  J{\'e}gou}]{lample2018word}
Alexis Conneau, Guillaume Lample, Marc'Aurelio Ranzato, Ludovic Denoyer, and
  Herv{\'e} J{\'e}gou. 2017.
\newblock Word translation without parallel data.
\newblock \emph{arXiv preprint arXiv:1710.04087}.

\bibitem[{Grave et~al.(2018)Grave, Bojanowski, Gupta, Joulin, and
  Mikolov}]{grave2018learning}
Edouard Grave, Piotr Bojanowski, Prakhar Gupta, Armand Joulin, and Tomas
  Mikolov. 2018.
\newblock Learning word vectors for 157 languages.
\newblock In \emph{Proceedings of the International Conference on Language
  Resources and Evaluation (LREC 2018)}.

\bibitem[{Heinzerling and Strube(2018)}]{heinzerling2018bpemb}
Benjamin Heinzerling and Michael Strube. 2018.
\newblock Bpemb: Tokenization-free pre-trained subword embeddings in 275
  languages.
\newblock In \emph{Proceedings of the Eleventh International Conference on
  Language Resources and Evaluation (LREC-2018)}.

\bibitem[{Kiela et~al.(2018)Kiela, Wang, and Cho}]{kiela2018dynamic}
Douwe Kiela, Changhan Wang, and Kyunghyun Cho. 2018.
\newblock Dynamic meta-embeddings for improved sentence representations.
\newblock In \emph{Proceedings of the 2018 Conference on Empirical Methods in
  Natural Language Processing}, pages 1466--1477.

\bibitem[{Lafferty et~al.(2001)Lafferty, McCallum, and
  Pereira}]{Lafferty2001ConditionalRF}
John~D. Lafferty, Andrew McCallum, and Fernando Pereira. 2001.
\newblock Conditional random fields: Probabilistic models for segmenting and
  labeling sequence data.
\newblock In \emph{ICML}.

\bibitem[{Lample et~al.(2016)Lample, Ballesteros, Subramanian, Kawakami, and
  Dyer}]{lample2016neural}
Guillaume Lample, Miguel Ballesteros, Sandeep Subramanian, Kazuya Kawakami, and
  Chris Dyer. 2016.
\newblock Neural architectures for named entity recognition.
\newblock In \emph{Proceedings of the 2016 Conference of the North American
  Chapter of the Association for Computational Linguistics: Human Language
  Technologies}, pages 260--270.

\bibitem[{Liu et~al.(2019)Liu, Xu, Winata, and
  Fung}]{liu-etal-2019-incorporating}
Zihan Liu, Yan Xu, Genta~Indra Winata, and Pascale Fung. 2019.
\newblock \href {https://www.aclweb.org/anthology/W19-5327} {Incorporating word
  and subword units in unsupervised machine translation using language model
  rescoring}.
\newblock In \emph{Proceedings of the Fourth Conference on Machine Translation
  (Volume 2: Shared Task Papers, Day 1)}, pages 275--282, Florence, Italy.
  Association for Computational Linguistics.

\bibitem[{Mikolov et~al.(2013)Mikolov, Sutskever, Chen, Corrado, and
  Dean}]{mikolov2013distributed}
Tomas Mikolov, Ilya Sutskever, Kai Chen, Greg~S Corrado, and Jeff Dean. 2013.
\newblock Distributed representations of words and phrases and their
  compositionality.
\newblock In \emph{Advances in neural information processing systems}, pages
  3111--3119.

\bibitem[{Murom{\"a}gi et~al.(2017)Murom{\"a}gi, Sirts, and
  Laur}]{muromagi2017linear}
Avo Murom{\"a}gi, Kairit Sirts, and Sven Laur. 2017.
\newblock Linear ensembles of word embedding models.
\newblock In \emph{Proceedings of the 21st Nordic Conference on Computational
  Linguistics}, pages 96--104.

\bibitem[{Pennington et~al.(2014)Pennington, Socher, and
  Manning}]{pennington2014glove}
Jeffrey Pennington, Richard Socher, and Christopher Manning. 2014.
\newblock Glove: Global vectors for word representation.
\newblock In \emph{Proceedings of the 2014 conference on empirical methods in
  natural language processing (EMNLP)}, pages 1532--1543.

\bibitem[{dos Santos and Zadrozny(2014)}]{Santos2014LearningCR}
C{\'i}cero~Nogueira dos Santos and Bianca Zadrozny. 2014.
\newblock Learning character-level representations for part-of-speech tagging.
\newblock In \emph{ICML}.

\bibitem[{Sennrich et~al.(2016)Sennrich, Haddow, and
  Birch}]{sennrich2016neural}
Rico Sennrich, Barry Haddow, and Alexandra Birch. 2016.
\newblock Neural machine translation of rare words with subword units.
\newblock In \emph{Proceedings of the 54th Annual Meeting of the Association
  for Computational Linguistics (Volume 1: Long Papers)}, volume~1, pages
  1715--1725.

\bibitem[{Trivedi et~al.(2018)Trivedi, Rangwani, and Singh}]{trivedi2018iit}
Shashwat Trivedi, Harsh Rangwani, and Anil~Kumar Singh. 2018.
\newblock Iit (bhu) submission for the acl shared task on named entity
  recognition on code-switched data.
\newblock In \emph{Proceedings of the Third Workshop on Computational
  Approaches to Linguistic Code-Switching}, pages 148--153.

\bibitem[{Vaswani et~al.(2017)Vaswani, Shazeer, Parmar, Uszkoreit, Jones,
  Gomez, Kaiser, and Polosukhin}]{vaswani2017attention}
Ashish Vaswani, Noam Shazeer, Niki Parmar, Jakob Uszkoreit, Llion Jones,
  Aidan~N Gomez, {\L}ukasz Kaiser, and Illia Polosukhin. 2017.
\newblock Attention is all you need.
\newblock In \emph{Advances in neural information processing systems}, pages
  5998--6008.

\bibitem[{Wang et~al.(2018)Wang, Cho, and Kiela}]{wang2018code}
Changhan Wang, Kyunghyun Cho, and Douwe Kiela. 2018.
\newblock Code-switched named entity recognition with embedding attention.
\newblock In \emph{Proceedings of the Third Workshop on Computational
  Approaches to Linguistic Code-Switching}, pages 154--158.

\bibitem[{Wieting et~al.(2016)Wieting, Bansal, Gimpel, and
  Livescu}]{wieting2016charagram}
John Wieting, Mohit Bansal, Kevin Gimpel, and Karen Livescu. 2016.
\newblock Charagram: Embedding words and sentences via character n-grams.
\newblock In \emph{Proceedings of the 2016 Conference on Empirical Methods in
  Natural Language Processing}, pages 1504--1515.

\bibitem[{Winata et~al.(2019)Winata, Lin, and Fung}]{winata2019learning}
Genta~Indra Winata, Zhaojiang Lin, and Pascale Fung. 2019.
\newblock Learning multilingual meta-embeddings for code-switching named entity
  recognition.
\newblock In \emph{Proceedings of the 4th Workshop on Representation Learning
  for NLP (RepL4NLP-2019)}, pages 181--186.

\bibitem[{Winata et~al.(2018{\natexlab{a}})Winata, Madotto, Wu, and
  Fung}]{winata2018code}
Genta~Indra Winata, Andrea Madotto, Chien-Sheng Wu, and Pascale Fung.
  2018{\natexlab{a}}.
\newblock Code-switching language modeling using syntax-aware multi-task
  learning.
\newblock In \emph{Proceedings of the Third Workshop on Computational
  Approaches to Linguistic Code-Switching}, pages 62--67.

\bibitem[{Winata et~al.(2018{\natexlab{b}})Winata, Wu, Madotto, and
  Fung}]{winata2018bilingual}
Genta~Indra Winata, Chien-Sheng Wu, Andrea Madotto, and Pascale Fung.
  2018{\natexlab{b}}.
\newblock Bilingual character representation for efficiently addressing
  out-of-vocabulary words in code-switching named entity recognition.
\newblock In \emph{Proceedings of the Third Workshop on Computational
  Approaches to Linguistic Code-Switching}, pages 110--114.

\bibitem[{Xu et~al.(2018)Xu, Madotto, Wu, Park, and Fung}]{xu2018emo2vec}
Peng Xu, Andrea Madotto, Chien-Sheng Wu, Ji~Ho Park, and Pascale Fung. 2018.
\newblock Emo2vec: Learning generalized emotion representation by multi-task
  training.
\newblock In \emph{Proceedings of the 9th Workshop on Computational Approaches
  to Subjectivity, Sentiment and Social Media Analysis}, pages 292--298.

\bibitem[{Yin and Sch{\"u}tze(2016)}]{yin2016learning}
Wenpeng Yin and Hinrich Sch{\"u}tze. 2016.
\newblock Learning word meta-embeddings.
\newblock In \emph{Proceedings of the 54th Annual Meeting of the Association
  for Computational Linguistics (Volume 1: Long Papers)}, volume~1, pages
  1351--1360.

\end{thebibliography}
\bibliographystyle{acl_natbib}

% \appendix

\end{document}